\pdfoutput=1

\documentclass[11pt]{article}

\usepackage[final]{acl}

\usepackage{times}
\usepackage{latexsym}

\usepackage[T1]{fontenc}

\usepackage[utf8]{inputenc}

\usepackage{microtype}

\usepackage{inconsolata}

\usepackage{graphicx}

\usepackage{amsmath}
\usepackage[linesnumbered,ruled,vlined]{algorithm2e}
\usepackage{booktabs}
\usepackage{multirow} 
\usepackage{array}
\usepackage{stfloats}

\title{Token-Level Marginalization for Multi-Label LLM Classifiers}

\author{
  Anjaneya Praharaj\thanks{Equal contribution.} \\
  ServiceNow, India \\
  \And
  Jaykumar Kasundra\footnotemark[1] \\
  ServiceNow, India \\
}

\begin{document}
\maketitle
\begin{abstract}
This paper addresses the critical challenge of deriving interpretable confidence scores from generative language models (LLMs) when applied to multi-label content safety classification. While models like LLaMA Guard are effective for identifying unsafe content and its categories, their generative architecture inherently lacks direct class-level probabilities, which hinders model confidence assessment and performance interpretation. This limitation complicates the setting of dynamic thresholds for content moderation and impedes fine-grained error analysis. This research proposes and evaluates three novel token-level probability estimation approaches to bridge this gap. The aim is to enhance model interpretability and accuracy, and evaluate the generalizability of this framework across different instruction-tuned models. Through extensive experimentation on a synthetically generated, rigorously annotated dataset, it is demonstrated that leveraging token logits significantly improves the interpretability and reliability of generative classifiers, enabling more nuanced content safety moderation. The code and the datasets are available \href{https://github.com/the-fenrir/MarginalProb4Classification}{here}.
\end{abstract}
\section{Introduction}

The rise of user-generated content has heightened the importance of content safety on digital platforms. Effective moderation systems must not only detect harmful content but also accurately categorize violations. Large Language Models (LLMs), known for their robust language understanding, are increasingly central to this task \cite{padhi2024granite, zeng2024shieldgemma, inan2023llama}. Models like LLaMA Guard \cite{inan2023llama} have been adapted for multi-label classification, producing structured outputs such as \texttt{`unsafe\textbackslash nS1, S3`}, aligned with a predefined safety taxonomy.

However, generative models like LLaMA Guard lack native support for producing confidence scores per predicted label, unlike discriminative classifiers. This absence complicates tasks such as thresholding, prioritization, and error analysis, which are critical in high-stakes settings \cite{geng-etal-2024-survey, 10.5555/3692070.3692492, tian2023just}. Without interpretable confidence, such systems risk both over-censorship and under-moderation.

To mitigate this, we introduce a framework that derives category-level confidence scores from token-level probabilities during autoregressive decoding. Following prior work \cite{cheng2024instruction, zhang2025token}, we evaluate three types of uncertainty estimation strategies: conditional probability, joint probability, and marginal probability.

\noindent \textbf{Contributions:}
\begin{enumerate}
    \item A principled method to extract confidence scores from generative LLMs;
    \item A comparison of multiple probability estimation techniques;
    \item Demonstration of the method’s generalizability to instruction-tuned models.
\end{enumerate}

\begin{figure*}[t]
  \centering
    \includegraphics[width=0.9\linewidth]{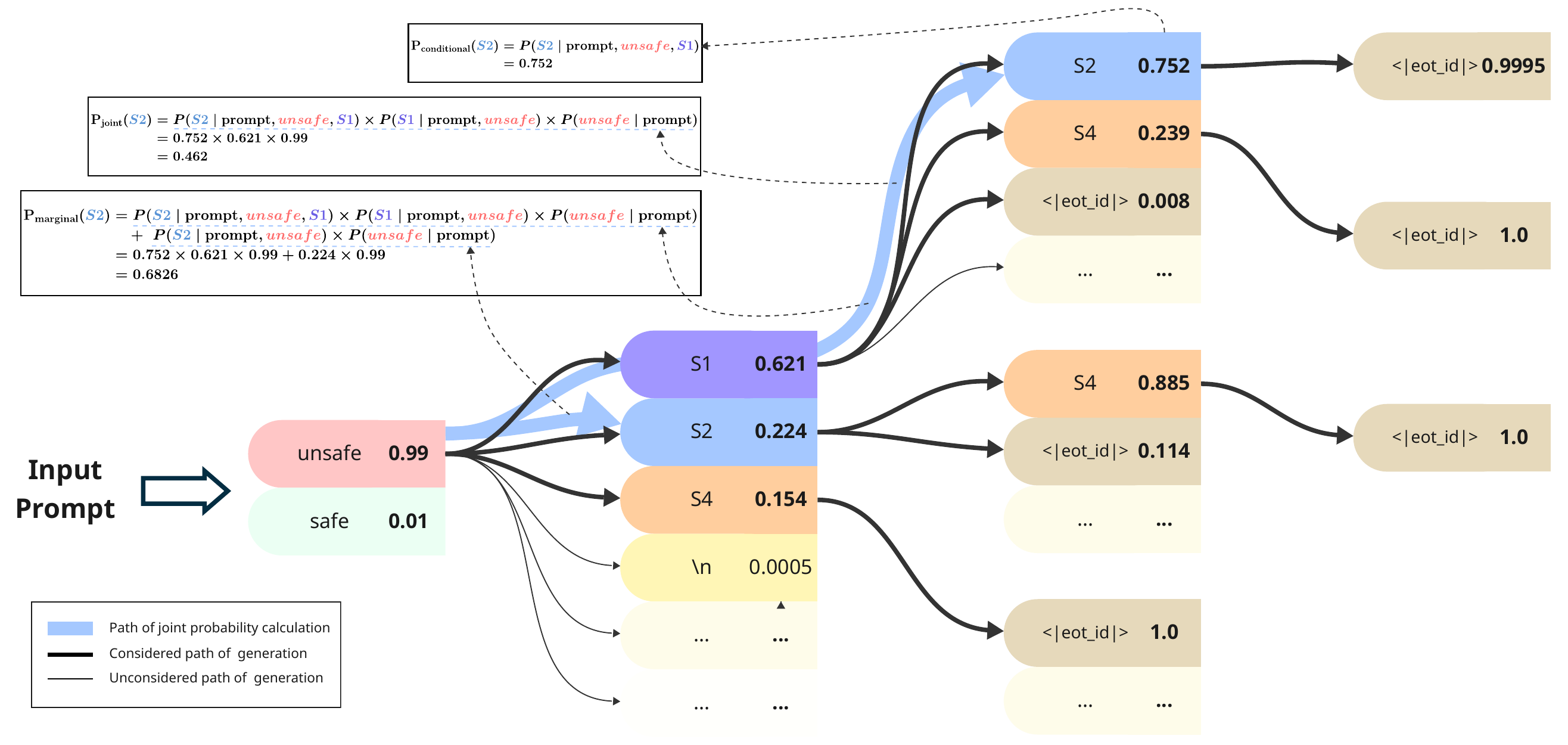} \hfill
    
    \label{fig:methods}
    \caption{We explore Conditional, Joint, and Marginal probability-based approaches to estimate model confidence. The category labels (e.g., S1, S3, etc.) correspond to classes defined in the LLaMA Guard taxonomy and are treated as tokens for simplicity.}
\end{figure*}

\section{Related Work}
Recent work has explored deriving confidence estimates from generative large language models (LLMs) \cite{ma2025estimating, xia2025survey, yang2024verbalized, 10.1162/tacl_a_00737}. Given their token-by-token decoding mechanism, researchers have proposed using logits and log-probabilities to estimate uncertainty \cite{mena2021survey, vazhentsev-etal-2025-token, yang-etal-2025-maqa}. Log-probabilities which are obtained via softmax over logits, can provide token or sequence-level likelihoods through methods such as joint or conditional probability aggregation \cite{fadeeva2024fact}. 

Methods like Logits-induced Token Uncertainty (LogU) compute token-level uncertainty efficiently without sampling, enabling applications in reranking and prompt engineering \cite{ma2025estimating}. Claim-conditioned probability estimation has been used to assess uncertainty around specific factual claims in tasks such as fact-checking \cite{fadeeva2024fact}, and prompt recovery techniques like \textsc{logit2prompt} leverage similar signal \cite{morris2023language}.

However, most prior work focuses on single-label classification or holistic sequence scoring. The challenge of systematically mapping token-level uncertainty to structured multi-label confidence remains underexplored.

\section{Methodology}
\subsection{Problem Formulation: Generative Models as Multi-label Classifiers}

Formally, let $X$ represent a textual content instance (input) and let $Y$ denote the set of $K$ predefined safety categories, $\mathcal{C} = \{C_1, C_2, \ldots, C_K\}$. For a multi-label classification task, an input instance $x$ can be associated with any subset of these labels, i.e., $y \subseteq \mathcal{C}$. This can be represented by a binary vector $y = [y_1, y_2, \ldots, y_K]$, where $y_i = 1$ if category $C_i$ is violated, and $y_i = 0$ otherwise.

Generative LLMs, such as LLaMA Guard, are trained to model the joint probability distribution of input and output tokens, $P(X, T)$, or, more commonly, the conditional probability of output tokens given the input, $P(T \mid X)$. Here, 

\[
T = (t_1, t_2, \ldots, t_L)
\]

represents the generated sequence of tokens that constitutes the classification output (e.g., \texttt{"unsafe\textbackslash nS1, S3"}). 

The fundamental challenge lies in deriving interpretable and reliable category-level confidence scores, $P(y_i = 1 \mid X)$, from this generative output, which is a sequence of tokens rather than explicit class probabilities.

\subsection{Token-Level Probability Estimation Approaches}
\label{estimation_approaches}

\begin{algorithm*}
\caption{Compute Marginal Probability of Label via Beam-like DFS with Max Token Cutoff}
\label{marginal_prob_code}
\DontPrintSemicolon
\SetAlgoLined
\SetKwFunction{FMain}{ComputeMarginalProbability}
\SetKwFunction{FDFS}{DFS}
\SetKwProg{Fn}{Function}{:}{}
\SetKwProg{Proc}{Procedure}{:}{}
\SetKw{KwInit}{Initialize}
\SetKw{KwGen}{Generate}
\SetKw{KwForEach}{For each}
\SetKw{KwIf}{If}
\SetKw{KwReturn}{Return}
\SetKw{KwBreak}{break}
\SetKw{KwContinue}{continue}
\SetKw{KwCall}{Call}

\Fn{\FMain{inputs, labels, max\_new\_tokens}}{
    \KwInit probabilities[label] $\gets$ 0 for each label\;
    
    \Proc{\FDFS{inputs, current\_probability, depth}}{
        \KwIf{current\_probability $<$ $1e^{-7}$}{
            \KwReturn
        }
        \KwGen next token logits using model\;
        top\_tokens $\gets$ Get top-$p$ tokens with their probabilities\;

        \KwForEach{(token, probability) in top\_tokens}{
            new\_inputs $\gets$ Append token to input\_ids and attention\_mask\;
            generation $\gets$ Decode new\_inputs to text\;

            \KwForEach{label in labels}{
                \KwIf{generation ends with label}{
                    probabilities[label] $\mathrel{+}= current\_probability \times probability$\;
                }
            }

            \KwIf{token is EOS and probability $\geq$ 0.7}{
                \KwBreak \tcp*{Stop exploring this path}
            }

            \KwIf{EOS token is among top tokens and this is the third token}{
                \KwBreak \tcp*{Stop exploring this path}
            }

            \KwIf{depth = max\_new\_tokens or token is EOS}{
                \KwContinue \tcp*{Skip recursion}
            }

            \KwCall \FDFS{new\_inputs, current\_probability $\times$ probability, depth + 1}\;
        }
    }

    \KwCall \FDFS{inputs, 1.0, 0}\;
    \KwReturn probabilities\;
}
\end{algorithm*}

All proposed methods leverage the raw, un-normalized scores (logits) generated by the LLM's final layer for each token in its vocabulary. These logits are then transformed into probabilities via a softmax function.

\subsubsection{Conditional Probability}

This approach computes the likelihood of a label token (e.g., \texttt{"S1"}) appearing at a specific step in the output, conditioned on the input prompt and previously generated tokens. It reflects the model's immediate probability of generating a given safety label during decoding.

For a target label $C_i$ represented by token(s) $t_{C_i}$, its conditional probability at generation step $j$ is:

\begin{equation}
\label{eq:conditional_prob_eq}
P(t_j = t_{C_i} \mid X, t_1, \ldots, t_{j-1}),
\end{equation}

which is obtained directly from the softmax output at step $j$.

In multi-label settings, we identify the label tokens (e.g., \texttt{"S1"}, \texttt{"S3"}) in the output and log their probabilities at generation. \footnote{When labels span multiple tokens (as in LLaMA Guard, where \texttt{"S1"} is tokenized as \texttt{'S'}, \texttt{'1'}), the probability of the final token (e.g., \texttt{'1'}) is used as a proxy for the label's likelihood.}

\subsubsection{Joint Probability}

This method computes the joint probability of generating each individual token in the output, conditioned on the input prompt and all previously generated tokens. For any target token $t_j$ in the generated sequence $T = (t_1, t_2, \ldots, t_L)$, the joint probability up to and including $t_j$ is given by:
\begin{multline}
\label{eq:joint_prob_eq}
P(t_{\leq j} \mid X) = P(t_1 \mid X) \times P(t_2 \mid X, t_1) \times \cdots \\
\times P(t_j \mid X, t_1, \ldots, t_{j-1})
\end{multline}
In practice, to improve numerical stability, the logarithm of the joint probability is computed as a sum of log probabilities.

\subsubsection{Marginal Probability}

Marginal probability estimates the overall likelihood of a specific label $C_i$ appearing in the model's output, considering all possible sequences containing that label, given an input $X$:

\begin{equation}
\label{eq:marginal_prob_eq}
P(C_i \mid X) = \sum_{T \in \mathcal{T}_{C_i}} P(T \mid X),
\end{equation}
where $\mathcal{T}_{C_i}$ denotes the set of output sequences that include $C_i$. The joint probability of each sequence $T = (t_1, \ldots, t_L)$ is given by:

\begin{equation}
\label{eq:joint_sequence_prob_eq}
P(T \mid X) = \prod_{j=1}^{L} P(t_j \mid X, t_{<j}).
\end{equation}

While theoretically comprehensive, capturing the true likelihood of label presence, this formulation is computationally intractable due to the exponential size of $\mathcal{T}_{C_i}$.

To approximate this, we adopt a constrained decoding strategy, detailed in Algorithm~\ref{marginal_prob_code}

\begin{enumerate}
    \item \textbf{Top-$p$ Filtering}: At each step, only tokens whose cumulative probability is below a threshold (e.g., 0.99) are considered, following nucleus sampling to prune unlikely paths.
    
    \item \textbf{Maximum generation depth}: We set a limit on the maximum number of tokens that can be generated along any given path.
    
    \item \textbf{Early Stopping on \texttt{[EOS]}}: Decoding halts upon generating an end-of-sequence token, ensuring that only complete outputs contribute to the final estimate.
\end{enumerate}

\begin{figure*}[htbp]
    \centering
    \includegraphics[width=0.95\linewidth]{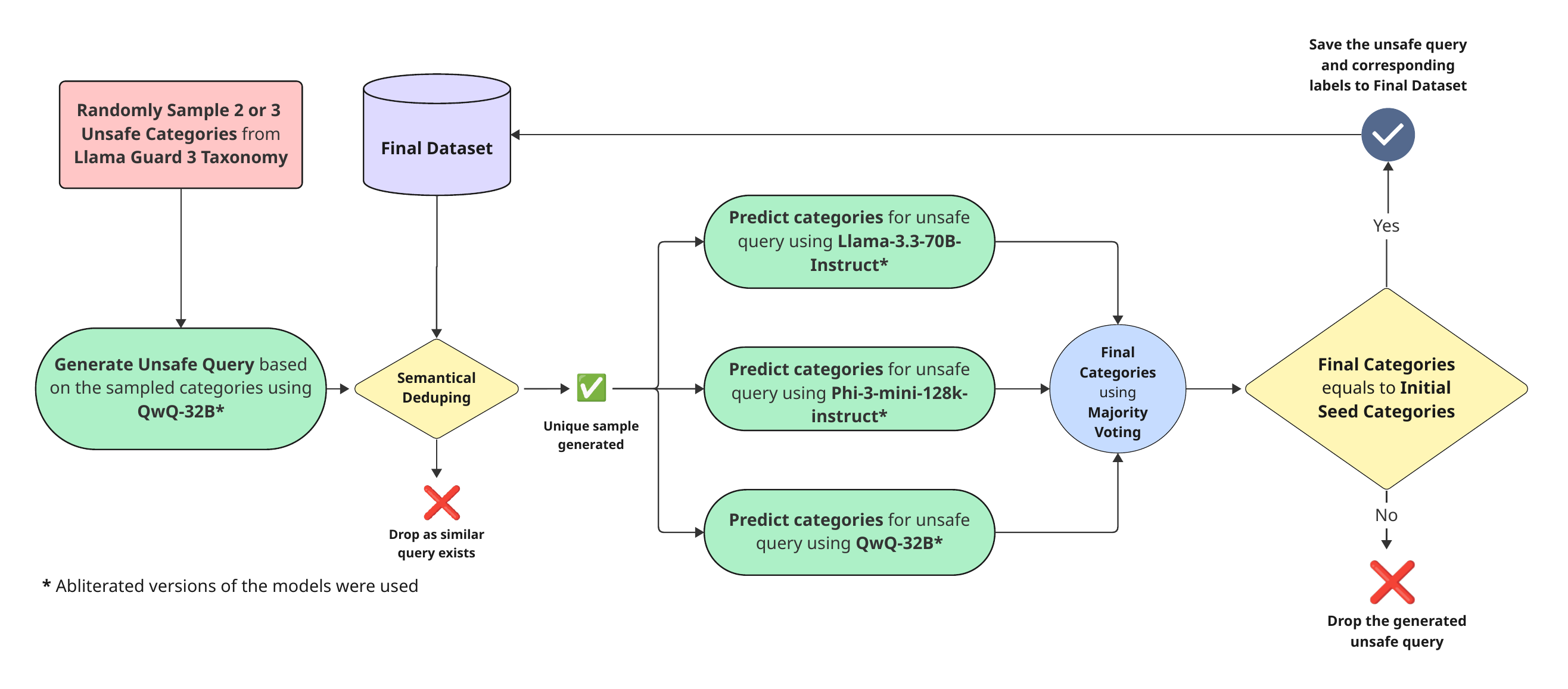} 
    \caption{An overview of the synthetic data generation pipeline used for generating the evaluation data. The models employed in this process include Qwen/QwQ-32B~\cite{qwq_huihui}, Meta-Llama/Llama-3.3-70B-Instruct~\cite{llama3_huihui}, and Microsoft/Phi-3-mini-128k-Instruct~\cite{dolphin_phi3}. Abliterated versions of these models were utilized to enable the generation of unsafe and offensive content.}
    \label{fig:data_gen_flow}
\end{figure*}

This approximation balances tractability with fidelity, enabling category-level marginal probability estimation in practice.

The distinction between conditional, joint, and marginal probabilities is crucial, as each offers a unique perspective on the model's confidence. Conditional probability focuses on the likelihood of individual label tokens at their point of generation. Joint probability assesses the confidence of the entire predicted label string. Marginal probability, being the most complex, attempts to capture the overall likelihood of a label independent of its exact position or co-occurrence with other specific tokens in the output string.

\subsection{Data Generation and Annotation}
As LLaMA Guard has not officially released any test datasets and no publicly available benchmark exists that aligns with its taxonomy, we opted to synthetically generate the evaluation data (Figure~\ref{fig:data_gen_flow}). Each content instance in the synthetic dataset is crafted to violate at least 2–3 safety categories based on LLaMA Guard 3 taxonomy. This controlled generation ensures a diverse set of multi-label examples, allowing for comprehensive evaluation across various safety categories. 

To ensure accurate ground truth labels, each data point's category annotations are derived by three separate LLMs. Only examples with at least 2 out of 3 model agreements matching the ground truth are retained. This reconciliation strategy creates a highly reliable "gold standard" dataset for evaluation. The final evaluation dataset consists of 2.3k records, with each category containing between 229 and 491 samples.

\begin{table*}[t]
\centering
\small  
\begin{tabular}{p{1.2cm} l c c c c}
\toprule
\multirow{2}{*}{\textbf{Model}} & \multirow{2}{*}{\textbf{Method}} & \multicolumn{2}{c}{\textbf{Synthetic Dataset}} & \multicolumn{2}{c}{\textbf{Beavertails}} \\
\cmidrule(lr){3-4} \cmidrule(lr){5-6}
& & \textbf{F1} & \textbf{AUCROC} & \textbf{F1} & \textbf{AUCROC} \\
\midrule
\multirow{7}{*}{\rotatebox{90}{\parbox{1.8cm}{\centering LLaMA Guard 2}}}
& Greedy Generation & 0.644 & -- & 0.430 & -- \\
& Probability Uncertainty & 0.496 & -- & 0.431 & -- \\
& Entropy Uncertainty & 0.496 & -- & 0.431 & -- \\
& LogTokU & 0.533 & -- & 0.430 & -- \\
& Conditional Probability & 0.644 & 0.756 & 0.430 & 0.649 \\
& Joint Probability & 0.644 & 0.754 & 0.430 & 0.649 \\
& Marginal Probability & \textbf{0.658} & \textbf{0.824} & \textbf{0.442} & \textbf{0.705} \\

\midrule
\multirow{7}{*}{\rotatebox{90}{\parbox{1.8cm}{\centering LLaMA Guard 3}}}
& Greedy Generation & 0.701 & -- & 0.418 & -- \\
& Probability Uncertainty & 0.698 & -- & 0.417 & -- \\
& Entropy Uncertainty & 0.698 & -- & 0.417 & -- \\
& LogTokU & 0.669 & -- & 0.418 & -- \\
& Conditional Probability & 0.701 & 0.777 & 0.418 & 0.640 \\
& Joint Probability & 0.700 & 0.777 & 0.419 & 0.640 \\
& Marginal Probability & \textbf{0.768} & \textbf{0.906} & \textbf{0.449} & \textbf{0.805} \\

\midrule
\multirow{7}{*}{\centering\rotatebox{90}{\parbox{2cm}{\centering LLaMA 3.1\\8B Instruct}}}
& Greedy Generation & 0.697 & -- & 0.373 & -- \\
& Probability Uncertainty & 0.453 & -- & 0.420 & -- \\
& Entropy Uncertainty & 0.453 & -- & 0.420 & -- \\
& LogTokU & 0.462 & -- & \textbf{0.426} & -- \\
& Conditional Probability & 0.704 & 0.876 & 0.376 & 0.677 \\
& Joint Probability & 0.626 & 0.874 & 0.401 & 0.678 \\
& Marginal Probability & \textbf{0.738} & \textbf{0.934} & 0.424 & \textbf{0.809} \\

\bottomrule
\end{tabular}
\caption{Comparison of various methods across multiple LLM-based safety classifiers. ↑ indicates higher-is-better metrics; ↓ indicates lower-is-better.}
\label{tab:marginal_probs_metrics}

\end{table*}

\section{Evaluation}

\subsection{Benchmarks}
We evaluate our approaches using greedy decoding across all models. For comparison, we include uncertainty estimation techniques introduced by \cite{ma2025estimating}, namely Probability Uncertainty, Entropy Uncertainty, and LogTokU. In addition to our synthetically generated dataset, we incorporate the Beavertails benchmark \cite{ji2023beavertails} to assess the performance of different methods under standardized evaluation settings.

\subsection{Evaluation Metrics}
\label{sec:metrics}
We evaluate model performance using standard metrics for multi-label classification: F1-score and AUCROC. F1-score is the harmonic mean of precision and recall. We report micro-averaged F1 across all labels to capture overall performance. AUCROC evaluates the model’s ability to distinguish between positive and negative classes. For multi-label settings, it is averaged over all labels.




\subsection{Results}
The primary models considered for content safety classification is the LLaMA Guard models. Its direct classification output (e.g., "unsafe\\nS1, S3") will serve as the baseline for performance comparison. 

The evaluation of the Conditional, Joint, and Marginal probability methods (outlined in Section~\ref{estimation_approaches}) for deriving category-level confidence scores on the LLaMA Guard model is shown in Table~\ref{tab:marginal_probs_metrics}. The results show that leveraging token logits for probability estimation significantly improves classification performance. The results shows that the Marginal Probability, leveraging its ability to aggregate probabilities across multiple paths, provides the most robust and accurate confidence scores, leading to superior overall classification performance.

\subsection{Generalizability of the approach}

To assess the transferability of the proposed strategy, LLaMA 3.1-8B-Instruct is considered. This model is an instruction-tuned LLM that has not been explicitly fine-tuned for content safety. The proposed probability-based decoding approach will be applied to the model and the performance of our approach is compared against the vanilla greedy decoding approach. Evaluation results in Table~\ref{tab:marginal_probs_metrics} show that even without explicit safety fine-tuning, a general instruction-tuned model, when used in the multi-label classification setting would show improved performance with marginal probability based approach.

\section{Conclusion}
This paper addressed the challenge of deriving interpretable confidence scores from generative LLMs for multi-label content safety classification. We proposed three token-level probability estimation methods—Conditional, Joint, and Marginal—to extract confidence scores from token logits. Experiments on a synthetic dataset show that these methods, especially the Marginal approach, significantly enhance classification accuracy. Overall, this work demonstrates that generative models can be adapted into reliable, interpretable multi-label classifiers, enabling broader use.

\subsection{Limitations and Future Work}

This work presents a novel approach for deriving confidence scores from generative LLMs in multi-label settings, but several limitations remain.

First, evaluations were performed on synthetic datasets. While useful for controlled experimentation, such data may not fully reflect the complexity and ambiguity of real-world harmful content, despite efforts to simulate realistic label distributions.

Second, the marginal probability estimation is approximate and does not explore the full space of generation paths. While tractable, this limits accuracy. Future work could investigate more efficient or principled marginal estimation techniques and examine how decoding strategies (e.g., beam width, top-p sampling) affect robustness.

Finally, the marginal probability method incurs token-level overhead due to multiple path explorations, which may hinder real-time applications. Practical deployment will require strategies to reduce this cost, such as adaptive path selection or approximation schemes.

\bibliography{custom}

\end{document}